\documentclass[10pt,twocolumn,letterpaper]{article}

\usepackage{iccv}
\usepackage{times}
\usepackage{epsfig}
\usepackage{graphicx}
\usepackage{amsmath}
\usepackage{amssymb}

\usepackage[breaklinks=true,bookmarks=false]{hyperref}

\iccvfinalcopy 

\def\iccvPaperID{****} 

\ificcvfinal\fi

\begin{document}

\title{Three Steps to Multimodal Trajectory Prediction: Modality Clustering, Classification and Synthesis}

\author{Jianhua Sun, Yuxuan Li, Hao-Shu Fang, Cewu Lu\footnotemark[4] \\
Shanghai Jiao Tong University, China\\
{\tt\small \{gothic,yuxuan\_li,lucewu\}@sjtu.edu.cn}
{\tt\small fhaoshu@gmail.com}
}

\maketitle
\ificcvfinal\thispagestyle{empty}\fi
\renewcommand{\thefootnote}{\fnsymbol{footnote}}
\footnotetext[4]{Cewu Lu is the corresponding author.}

\begin{abstract}
Multimodal prediction results are essential for trajectory prediction task as there is no single correct answer for the future. Previous frameworks can be divided into three categories: regression, generation and classification frameworks. However, these frameworks have weaknesses in different aspects so that they cannot model the multimodal prediction task comprehensively. In this paper, we present a novel insight along with a brand-new prediction framework by formulating multimodal prediction into three steps: modality clustering, classification and synthesis, and address the shortcomings of earlier frameworks. Exhaustive experiments on popular benchmarks have demonstrated that our proposed method surpasses state-of-the-art works even without introducing social and map information. Specifically, we achieve $19.2\%$ and $20.8\%$ improvement on ADE and FDE respectively on ETH/UCY dataset. Our code will be made publicly available.
\end{abstract}

\section{Introduction}

Trajectory prediction~\cite{alahi2016social,gupta2018social,vemula2018social,sadeghian2019sophie,Lee_2017_CVPR,ivanovic2019trajectron,liang2019peeking,yu2020spatio,mangalam2020not,fang2020tpnet} is one of the cornerstones of autonomous driving and robot navigation, which investigates reasonable future states of traffic agents for the following decision-making process. Considering the uncertainty of human behaviors and the multimodal nature of the future~\cite{gupta2018social,Lee_2017_CVPR}, one great challenge of trajectory prediction lies in predicting all possible future trajectories of high probabilities.

\begin{figure}[t]
\begin{center}
   \includegraphics[width=\linewidth]{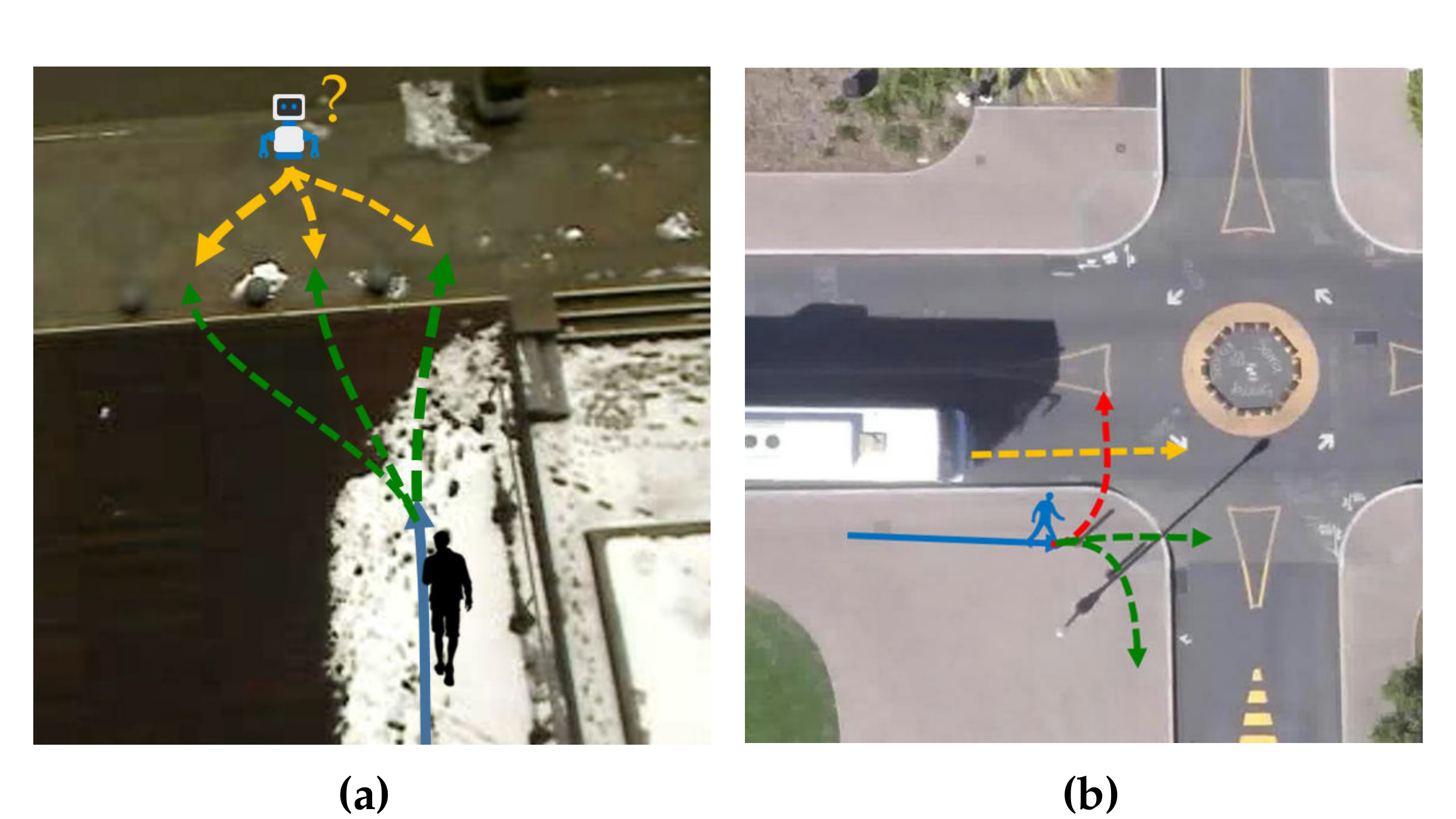}
\end{center}
\vspace{-5pt}
   \caption{Examples of why probability and determinacy of a prediction algorithm is important, where (a) illustrates a confusing case without probabilistic predictions and (b) illustrates a safety hazard of stochastic predictions. In figure (b), the red path should always be predicted for safety purposes. A detailed discussion is in Sec.~\ref{sec:related}.}
\vspace{-10pt}
\label{fig:badcase}
\end{figure}

To tackle this problem, previous research mainly follows three lines. The first line adds extra randomness for regression frameworks~\cite{yu2020spatio,liang2019peeking}, while the second generation line~\cite{gupta2018social,Lee_2017_CVPR,mangalam2020not,sadeghian2019sophie} models the multimodal nature by learning a distribution of the future. But both lines have two defects as shown in Fig.~\ref{fig:badcase}: i) lack of probability corresponding to each modality which may leave the decision process confusing, and ii) the prediction results are not deterministic which may leave potential safety risks. 

The third line of classification frameworks~\cite{chai2019multipath,phan2020covernet} gets rid of the two defects by classifying the observation to predefined future trajectories, as the classification operation can give probabilities and ensure determinacy. However, the classification frameworks still face certain weaknesses. First, the predefined trajectories are obtained by hand-crafted principles, thus it is difficult to capture comprehensive representations for future behaviors. Second, the predicted deterministic trajectories will be the same for different inputs classified to a same class, which fails to explore fine-grained motions for traffic agents deterministically. Due to these weaknesses, the performance of a classification framework lags behind state-of-the-art regression and generation models. Further, a highly annotated scene raster is required as the input of classifier which is difficult to access in many cases.

In this paper, we aim to explore a distinct formulation for trajectory prediction framework to address the shortcomings discussed above. We present the insight of \textbf{Prediction via modality Clustering, Classification and Synthesis} (PCCSNet) by solving multimodal prediction with a \textit{classification-regression} approach. In our vision, the modalities of the future are usually centralized around a few different behaviors which can be revealed by a series of learned modality representations. We can apply a deep clustering process on training samples and each center of clusters could represent a modality. Naturally, such a modality can be formulated into a class, and a classification network can be adopted to distinguish and score the modalities according to the observed trajectory in this manner. Finally, a synthesis process is used to regress prediction results for highly probable modalities with historical states and the modality representations. 

\begin{figure*}[tb!]
\vspace{-20pt}
\centering
\includegraphics[width=1.0\textwidth]{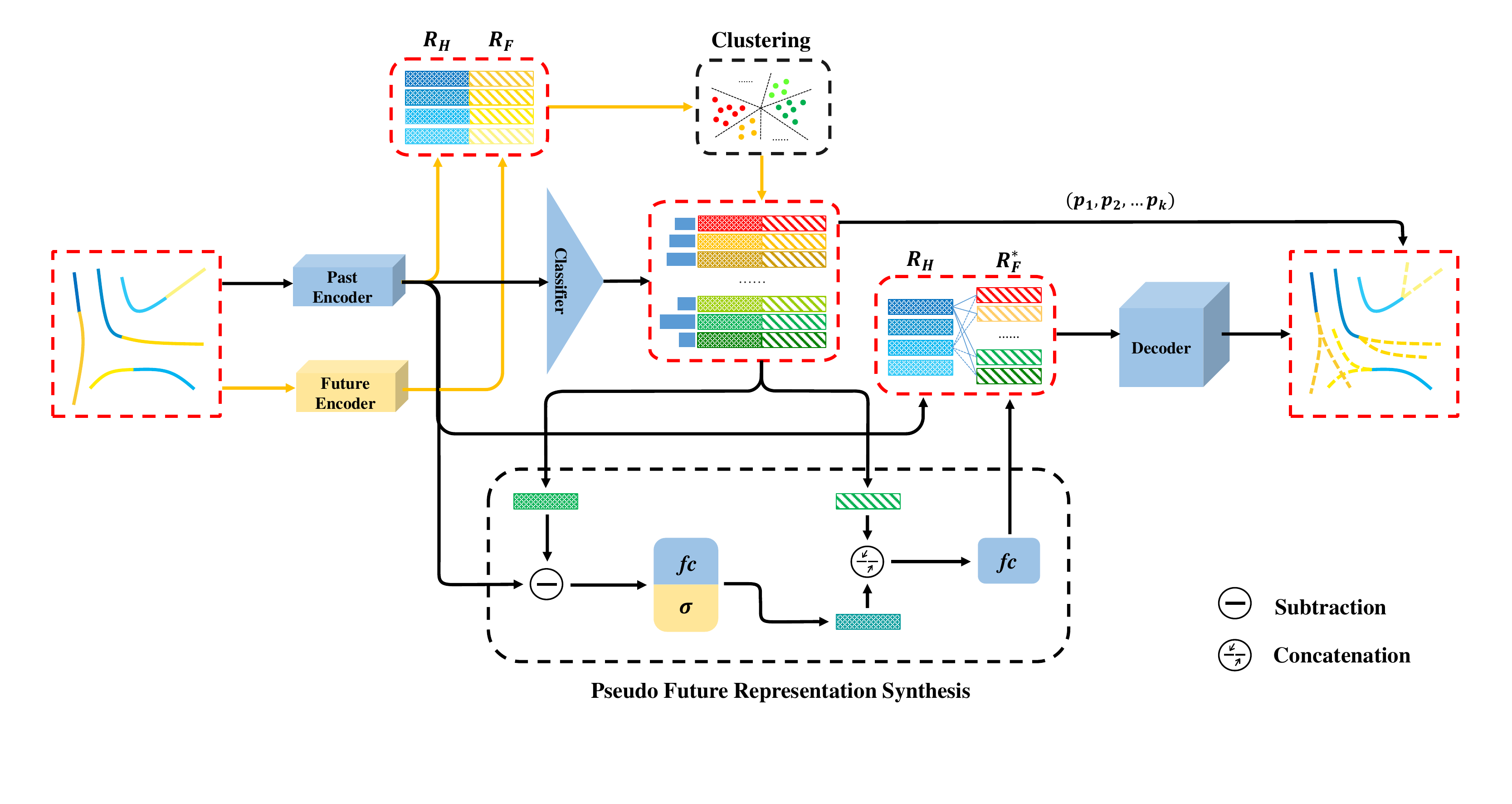}
\vspace{-40pt}
\caption{Overview of our proposed prediction pipeline. The arrows in yellow are only present in training, when both historical and future representations are concatenated and clustered for future usage. At test time, historical representations are fed into a classifier to score different modalities. The modality representations are then processed with the historical representations to synthesize multimodal predictions.}
\label{fig:network}
\vspace{-15pt}
\end{figure*}

We propose a modular designed framework to model this novel insight summarized in Fig.~\ref{fig:network}. States of agents are first fed into feature encoders to get deep historical and future representations for better clustering, classification and synthesis~\cite{bengio2013better}. These deep features are clustered for modality representations and used to train a classifier where the cluster assignments are seen as pseudo-labels. The classifier will score the modalities according to historical representations in test phase for probabilistic prediction. A synthesis module is then introduced to regress pseudo future representations for each modality, and finally both historical and synthesized future representations are decoded to get fine-grained deterministic predictions. Moreover, we newly propose a Modality Loss to enhance the capability of the classifier to identify multiple highly probable future modalities.

We conduct exhaustive experiments on multiple popular trajectory prediction benchmarks. In these experiments, our novel prediction framework exhibits high accuracy, great robustness and adequate projections for the future. Specifically, we achieve $19.2\%$ and $20.8\%$ improvement in average on ADE and FDE respectively on ETH~\cite{pellegrini2009you}/UCY~\cite{leal2014learning} datasets comparing with state-of-the-art method~\cite{yu2020spatio}. 

\section{Background and Related Work}
\label{sec:related}

The trajectory prediction task is proposed to forecast possible trajectories of an agent. It takes advantages from tracking~\cite{sadeghian2017tracking, pang2020tubetk} and human interactions~\cite{zhou2012understanding,yamaguchi2011you,li2020pastanet}, and has many applications in the field of robotics and autonomous driving~\cite{hirakawa2018survey, robicquet2016learning, wang2021structured, wang2020active}. Observing the multimodal nature of the future which can be interpreted as no single correct answer for the future~\cite{gupta2018social}, an important point lies in how to predict multiple highly probable trajectories. This task is named as multimodal trajectory prediction. Note that a small part of methods~\cite{Choi_2019_ICCV,liang2020garden} re-formulate this task by predicting probabilistic maps in pixel level. We mainly discuss prevailing approaches that outputs multiple possible trajectories of spatial coordinate system (meters) in real world in this paper.

Multimodal prediction task is non-trivial as a single input may map to multiple outputs. Early works \cite{alahi2016social, mehran2009abnormal, Xie_2013_ICCV, kitani2012activity} ignore the multimodality of the future and only aim at predicting the most possible future trajectory. Recently, a great number of research proposes various frameworks to formulate this non-functional relationship. They mainly follow three common practices, regression, generation and classification.

\vspace{5pt}\noindent\textbf{Regression Frameworks.} Regression models~\cite{alahi2016social,ma2019trafficpredict} are first proposed to solve unimodal prediction tasks and show great performance. However, these encode-decode structures are not able to give multimodal predictions, and some methods address this defect by adding noise~\cite{yu2020spatio} or using random initialization~\cite{liang2019peeking}. Although multiple different predictions can be obtained by imposing randomness on the model, it is difficult for randomness to accurately model the multimodal nature of future. 

\vspace{5pt}\noindent\textbf{Generation Frameworks.} Some research considers the multimodality of the future as a distribution, formulates trajectory prediction as a distribution fitting and sampling problem, and introduces generative models to solve it. DESIRE~\cite{Lee_2017_CVPR} first introduces stochastic model to learn the distribution of future states, and generates diverse predictions by sampling plausible hypotheses from that distribution. Following this formulation, plenty of  research~\cite{gupta2018social,mangalam2020not,sadeghian2019sophie} aims at designing different generative structures to pursue more reasonable outcomes and achieve the state-of-the-art performance. 

\vspace{5pt}\noindent\textbf{Classification Frameworks.} Some research~\cite{chai2019multipath, phan2020covernet} attempts to use a classification network to solve this problem by classifying on predefined artificial modalities. Multipath~\cite{chai2019multipath} clusters a fixed set of anchor trajectories with mean square error distance, and classifies the input to these anchors. CoverNet~\cite{phan2020covernet} revises Multipath by manually designing anchors. Approaches under this framework face three main weakness. First, the predefined trajectories are obtained by subjectively designated clustering distance or manually designed anchors, thus it is difficult to capture the full range of future behaviors. Second, it is hard for these predefined trajectories to capture fine-grained motions. Further, both methods require a highly annotated scene raster as input for classification which is difficult to access in many cases.

\vspace{5pt}\noindent\textbf{Probability and Determinacy.} The properties of probability and determinacy  are important for a multimodal prediction approach. i) Probability. The probabilistic property of future~\cite{joyce2003bayes} are extremely helpful to improve the effectiveness of the ensuing decision-making process. In Fig.~\ref{fig:badcase} (a), a probabilistic prediction can tell that the pedestrian is most unlikely to take the left path (on the reader's side). Therefore, the robot can follow the left path (on the reader's side) to avoid collision as far as possible. ii) Determinacy. A stochastic framework may leave huge potential safety risks. Fig.~\ref{fig:badcase} (b) gives a common case that a bus (autonomous vehicle) perceives a pedestrian at the crossroads when moving forward. A stochastic model cannot be proved to predict the trajectory in red every time. And if it fails, a traffic accident will happen. But most regression and generation models miss these two points, as equiprobable randomness is introduced. Previous approaches with classification frameworks can give probabilistic and deterministic predictions, but these frameworks fail in exploiting deep behavior representations and fine-grained motions as discussed above. 

We propose a probabilistic and deterministic framework in this work which can still capture deep behavior representations and give predictions at fine-grained level. The great differences between our proposed framework and previous classification frameworks will be discussed in Sec.~\ref{sec:discussion}.

\section{Approach}

In this section, we introduce a newly proposed PCCSNet prediction pipeline, which is illustrated in Fig.~\ref{fig:network}. Our main insight is to formulate the multimodal prediction framework as a classification-regression process. It is designed to model multimodal trajectory task more comprehensively by addressing the shortcomings of prior frameworks.

\subsection{Problem Definition}

Following previous works \cite{alahi2016social, gupta2018social}, we assume that each video is preprocessed by detection and tracking algorithms to obtain the spatial coordinates for each person at each timestep. We take the coordinate sequences $X$ in time step $[1,T_{obs}]$ as input, and predict top $k$ multiple reasonable coordinate sequences $\hat{\mathbb{Y}} = \{\hat{Y}_1, \hat{Y}_2, \dots, \hat{Y}_k\}$ in $[T_{obs+1}, T_{obs+pred}]$ along with their probabilities $\mathbb{P}=\{p_1, p_2, \dots, p_k\}$ as output. 

\subsection{Overview}

In PCCSNet, we introduce an intermediate variable named modality representation $M$ to formulate multimodal prediction framework into three steps of deep clustering, classification and synthesis. All of the possible modality representations can be obtained by clustering deep historical representations $R_H$ and future representations $R_F$ of training samples and gathered into a modality set $\mathbb{M}$
\begin{equation}
    R_H = f_H(X), R_F = f_F(Y)
    \label{eqn:encoding}
\end{equation}
\begin{equation}
\begin{split}
    \mathbb{M} =& clustering(\{[R_H^i, R_F^i]|i \in trainset\}) \\
        =& \{M_1, M_2, \dots, M_n\} 
    \label{eqn:modality set}
\end{split}
\end{equation}
where past encoder $f_H(\cdot)$ and future encoder $f_F(\cdot)$ are trained to learn better representations of historical trajectory $X$ and future trajectory $Y$ following \cite{bengio2013better}. 

Then, through modality classification and modality synthesis, we can acquire $\mathbb{\hat{Y}}$ along with its probability $\mathbb{P}$ by
\begin{equation}
    \mathbb{P} = g_\mathbb{M}(R_H)
    \label{eqn:classifier}
\end{equation}
\begin{equation}
    \hat{\mathbb{Y}} = \{\hat{Y}_i = h([R_H,M_i])|i \in [1,n]\}
    \label{eqn:synthesis}
\end{equation}
where $g_\mathbb{M}(\cdot)$ represents modality classification on $\mathbb{M}$ and $h(\cdot)$ represents modality synthesis. In this manner, we can predict probabilistic multimodal future trajectories deterministically. Note that we often predict $k$ ($k<n$) future paths with the top probabilities in practical terms to reduce the test time. 

In the following sections, we will introduce how we cluster and train the classification network $g_\mathbb{M}(\cdot)$ in Sec.~\ref{sec:Classification}. In Sec.~\ref{sec:Modality Loss}, we propose a novel Modality Loss to encourage the classifier to recognize multiple reasonable futures comprehensively instead of just the most likely one. Finally, we show how to synthesize one prediction result by $h(\cdot)$ in Sec.~\ref{sec:Prediction}.

\subsection{Classification with Modality Clustering}
\label{sec:Classification}

Following Eq.~\ref{eqn:modality set} and Eq.~\ref{eqn:classifier}, we need to construct modality set $\mathbb{M}$ and corresponding classifier $g_\mathbb{M}(\cdot)$.

\vspace{5pt}\noindent\textbf{Feature Encoder.} To capture better representations for deep clustering, classification and further synthesis procedure, we first encode the historical and future trajectories for each agent. Given that a trajectory is a time series and has a strong dependency and consistency between each time step according to \cite{sun2020recursive}, we adopt the BiLSTM architecture as our feature encoders. 

\vspace{5pt}\noindent\textbf{Clustering.} We believe that each modality of trajectory indicates behaviors and movements of the same kind, and in turn, we can express a modality representation $M$ with the average of a series of deep trajectory features, which can be written as
\begin{equation}
    M = AVG(\{[R_H^i, R_F^i]|i\in C\})
    \label{eqn:modality tmp}
\end{equation}
where $C$ is a cluster represented by trajectory ids. In our implementation, we use the clustering center of $C$ to represent the $AVG$ operation, and Eq.~\ref{eqn:modality tmp} can be rewritten as
\begin{equation}
    M = [R_H^c, R_F^c]
    \label{eqn:modality}
\end{equation}
where $R_H^c$ and $R_F^c$ are the values of historical and future representations in the clustering center. In this way, we create a bridge between modality construction and clustering. 

To generate distinct $C$s, we introduce a clustering algorithm. Considering the definition of $M$, we use $[R_H^i, R_F^i]$ as features for path id $i$ and a weighted L2 distance for clustering. Specifically, the distance is written as
\begin{equation}
    \mathcal{D} = w_H||R_H^1-R_H^2||_2 + w_F||R_F^1-R_F^2||_2
    \label{eqn:L2 distance}
\end{equation}
where $w_H$ and $w_F$ represent the weight for historical representations and future representations respectively.

Here we assume that the mean value and distribution of the training set are similar as those of the test set, which is usually how it works. The modality set $\mathbb{M}$ we construct will very nicely cover different kinds of future possibilities for test samples, as shown in Fig.~\ref{fig:clusters experiment}. 

\vspace{5pt}\noindent\textbf{Classifier.} Eq.~\ref{eqn:classifier} illustrates the functionality of our classifier. $g_\mathbb{M}(\cdot)$ receives the encoded feature $R_H$ of input trajectory $X$ and outputs possibilities for each modality. In our implementation, $g_\mathbb{M}(\cdot)$ is a three-layer MLP (Multilayer Perceptron) with a $tanh$ activation. We train the classifier by treating the cluster assignments as pseudo-labels.

\begin{figure}[t]
\begin{center}
   \includegraphics[width=0.95\linewidth]{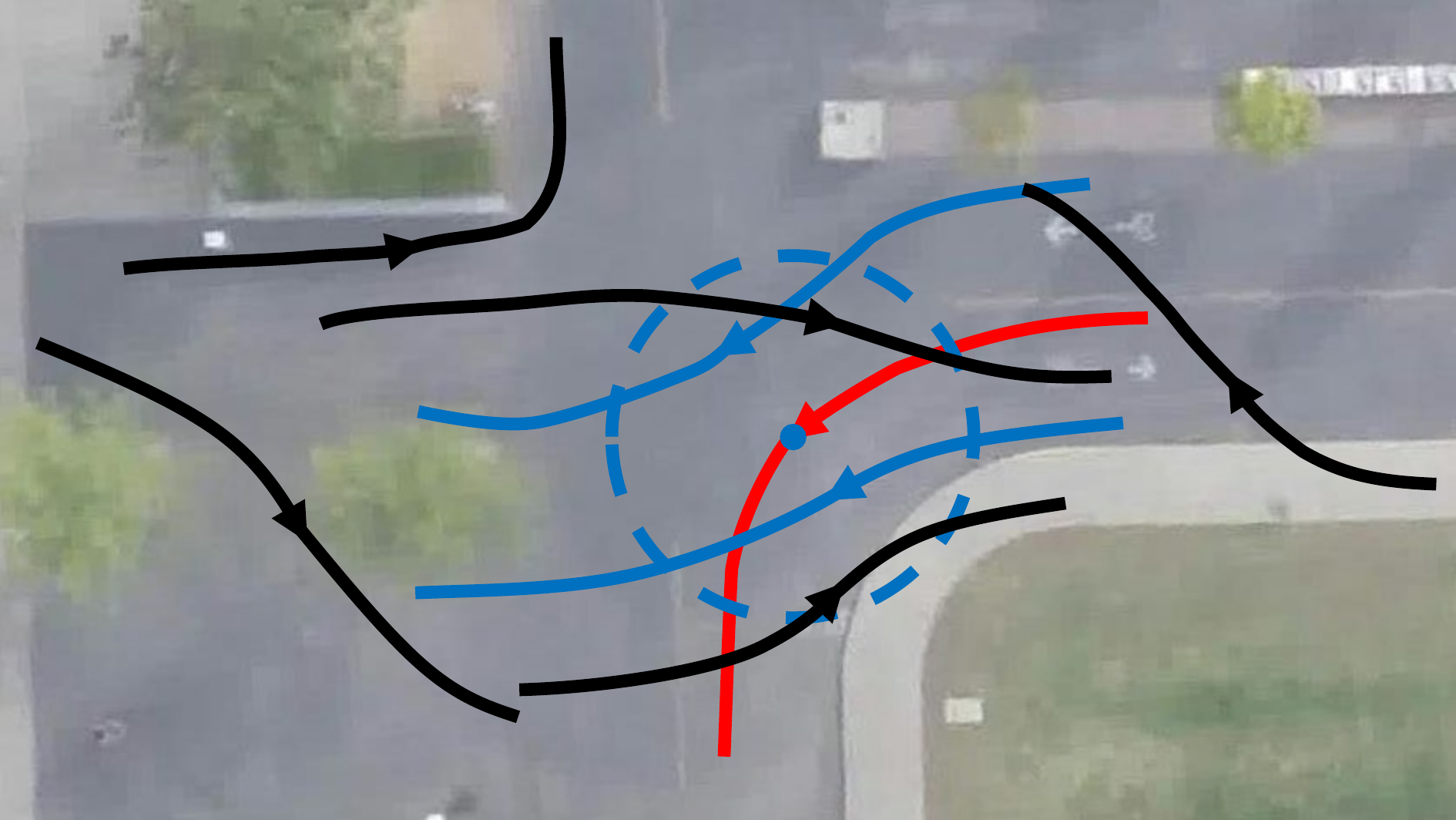}
\end{center}
\vspace{-5pt}
   \caption{A schematic on how to estimate other reasonable modalities (in blue) for a target path (in red).}
\label{fig:modality loss}
\vspace{-10pt}
\end{figure}

\subsection{Enhancing Diversity with Modality Loss}
\label{sec:Modality Loss}

Samples in traditional classification tasks have only one-class assignments (ground truth). This goes against the optimization goal of a multimodal classification task that the classifier should figure out a series of futures with high probabilities to happen. In this regard, we propose a statistical method to estimate reasonable and feasible pseudo future modalities for a target path. 

Specifically, Fig.~\ref{fig:modality loss} illustrates an example, where the red curve denotes the target path. We draw a circle $o$ centered on the end of its observed part with a radius $r$ and count other trajectories truncated by this circle in the entire time period of this scene. Then we group trajectories that have similar speeds and directions as the ground truth path. Qualified paths are highlighted in blue, and can be seen as other potential movements $Y^*_1, Y^*_2, \dots, Y^*_N$.

Then, the pseudo possibility of each modality is calculated according to these potential movements by
\begin{equation}
    p^*_j = \frac{|\{Y^*_i|Y^*_i \in M_j\}|}{N}
    \label{eqn:M possibility}
\end{equation}
and we define Modality Loss as
\begin{equation}
    \mathcal{L}_M = CrossEntropy(\mathbb{P},\mathbb{P}^*)
    \label{eqn:Modality Loss}
\end{equation}
where $\mathbb{P}$ denotes the classification result acquired from Eq.~\ref{eqn:classifier} and $\mathbb{P}^*$ denotes the pseudo label acquired from Eq.~\ref{eqn:M possibility}. In this way, we can use these statistical probabilities to supervise the classifier, which brings great diversity to our model. 

\subsection{Prediction with Modality Synthesis}
\label{sec:Prediction}

So far, we have calculated each modality $M$ and historical representation $R_H$ for an input observed path. In this section, we will discuss how $h(\cdot)$ in Eq.~\ref{eqn:synthesis} synthesizes a prediction result for each modality. 

\vspace{5pt}\noindent\textbf{Pseudo Future Representation for Each Modality.} In order to synthesize a future trajectory in line with a given modality, we need a corresponding pseudo future representation to indicate the future trajectory propensity. We propose a regression model to fit a pseudo future representation $R_F^*$ centered on $M$ for given $R_H$. 

A diagram of the regression is shown at the bottom of Fig.~\ref{fig:network}. We first subtract $R_H^c$ from $R_H$ and use an MLP with a \textit{sigmoid} activation to encode the difference. After that, the encoded features are concatenated with corresponding $R_F^c$ to form a new vector, which indicates the average of the future propensity as well as its bias to the input trajectory. Finally, the vector is fed into a fully connected layer to extract the pseudo future representation $R_F^*$. Each $R_F^*$ summarizes a behavior of modality $M$ and is able to reflect the tendency of future trajectory.

For training, we only compute $R_F^*$ of the cluster which input $R_H$ is assigned to. Corresponding $R_F$ and L2 loss are used to supervise the generation of $R_F^*$. 

\vspace{5pt}\noindent\textbf{Decoder.} With historical representation $R_H$ and pseudo future representation $R_F^*$ of different modalities, we can synthesize future trajectories using an LSTM decoder. We input $h_0 = [R_H, R_{F_i}^*]$ and output predicted $\hat{Y}_i$ for modality $M_i$. All modalities share the same parameters and we use exponential L2 Loss in \cite{sun2020recursive} for better performance. 

\subsection{Implementation Details}
In our implementation, both BiLSTM encoders have a hidden size of 48, while the LSTM decoder has a hidden size of 96. We choose the classic K-Means~\cite{macqueen1967some} algorithm for clustering where the hyper-parameter $K$ is set to $200$. For weight coefficients in Eq.~\ref{eqn:L2 distance}, we let $w_H=w_F=0.5$. To find the qualified paths for Modality Loss as shown in Fig.~\ref{fig:modality loss}, we set the radius $r=1$, and employ a $10\%$ limit on speed differences $\Delta v$ and a $0.1\pi$ limit on direction differences $\Delta \theta$.

\subsection{Discussion}
\label{sec:discussion}

Comparing with previous classification framework MultiPath~\cite{chai2019multipath}, our insight and approach show great differences. i) We encode deep representations for behaviors and cluster modalities on them. In our vision, human behaviors are too complicated to be represented by simple coordinate series~\cite{chai2019multipath}. This deep clustering process can explore much better representations for modalities. Further, the idea of metric learning is implied in this process while human behavior is rather complex to be clustered by manually designed distance~\cite{chai2019multipath}. ii) Our classifier does not require extra scene raster as input. Actually, our framework can work well with only historical paths as input and thus our approach has a strong ability to be generalized into most prediction cases. iii) A synthesis step is proposed to provide deterministically predicted trajectories at fine-grained level for each modality, while the deterministic trajectories will be the same for inputs classified to a same class in \cite{chai2019multipath} and the prediction space will be severely constrained.

We only introduce historical paths as past features in this paper for clarity since the highlight of this research is prediction framework rather than `social or contextual information'. However, it would be easy to incorporate other information into our proposed framework. Similar to Past Encoder, one can use a social/map encoder to encode these features and concatenate them with $R_H$.

\section{Experiment}

\subsection{Datasets}

Performance of our method is evaluated on popular datasets, including ETH~\cite{pellegrini2009you}/UCY~\cite{leal2014learning} Dataset and Stanford Drone Dataset~\cite{robicquet2016learning}. The ETH/UCY dataset is widely used for trajectory prediction benchmark~\cite{alahi2016social, gupta2018social, liang2019peeking, sadeghian2019sophie}, which consists five different sub-datasets (ETH, HOTEL, UNIV, ZARA1 and ZARA2). The Stanford Drone Dataset is a large-scale dataset including various agents. These trajectories are recorded by drone cameras in bird's eye view with sufficient diversity. 

In our experiments, we follow the same data preprocessing procedure and evaluation configuration as previous work~\cite{yu2020spatio,mangalam2020not}. To evaluate the accuracy of our prediction results, we use Average Displacement Error (ADE) and Final Displacement Error (FDE) as metrics. we observe historical trajectories for 3.2sec (8 frames) and predict future trajectories for 4.8sec (12 frames) at a frame rate of 0.4. 20 samples of the future trajectories are predicted. 

\begin{table*}[tb!]
\begin{center}
 \begin{tabular}{c|c||c|c||c|c|c||c}
 \hline
  Method & Input & ETH & HOTEL & UNIV & ZARA1 & ZARA2 & AVG \\
  \hline
    SGAN~\cite{gupta2018social} & P+S & 0.81 / 1.52 & 0.72 / 1.61 & 0.60 / 1.26 & 0.34 / 0.69 & 0.42 / 0.84 & 0.58  /1.18 \\
    Sophie~\cite{sadeghian2019sophie} & P+S+M & 0.70 / 1.43 & 0.76 / 1.67 & 0.54 / 1.24 & 0.30 / 0.63 & 0.38 / 0.78 & 0.54 / 1.15 \\
    Next~\cite{liang2019peeking} & P+M & 0.73 / 1.65 & 0.30 / 0.59 & 0.60 / 1.27 & 0.38 / 0.81 & 0.31 / 0.68 & 0.46 / 1.00 \\
    Social STGCNN~\cite{mohamed2020social} & P+S & 0.64 / 1.11 & 0.49 / 0.85 & 0.44 / 0.79 & 0.34 / 0.53 & 0.30 / 0.48 & 0.44 / 0.75 \\
    PECNet~\cite{mangalam2020not} & P+S & 0.54 / 0.87 & 0.18 / 0.24 & 0.35 / 0.60 & 0.22 / \textbf{0.39} & 0.17 / \textbf{0.30} & 0.29 / 0.48 \\
    STAR~\cite{yu2020spatio} & P+S & 0.36 / 0.65 & 0.17 / 0.36 & 0.31 / 0.62 & 0.26 / 0.55 & 0.22 / 0.46 & 0.26 / 0.53 \\
  \hline
    PCCSNet & P & \textbf{0.28} / \textbf{0.54} & \textbf{0.11} / \textbf{0.19} & \textbf{0.29} / \textbf{0.60} & \textbf{0.21} / 0.44 &  \textbf{0.15} / 0.34 & \textbf{0.21} / \textbf{0.42} \\
  \hline
 \end{tabular}
 \vspace{5pt}
 \caption{Comparison on ETH and UCY dataset for $T_{obs}=8$ and $T_{pred}=12$ (ADE/FDE), including SOTA STAR and PECNet. P denotes historical path, S denotes social information and M denotes map information. \cite{liang2019peeking} also uses pose information from AlphaPose~\cite{fang2017rmpe,li2019crowdpose}. It is worth mentioning that our approach outperforms other methods in average without using social and map information.}
\label{tab:eth}
\vspace{-15pt}
\end{center}
\end{table*}

\begin{table}[tb!]\footnotesize
\begin{center}
 \begin{tabular}{c||c|c|c||c}
 \hline
  Method & SGAN~\cite{gupta2018social} & Sophie~\cite{sadeghian2019sophie} & PECNet~\cite{mangalam2020not} & PCCSNet \\
  \hline
    Input & P+S & P+S & P+S & P \\
  \hline
    ADE & 27.23 & 16.27 & 9.96 & \textbf{8.62}  \\
    FDE & 41.44 & 29.38 & \textbf{15.88} & 16.16  \\
  \hline
 \end{tabular}
 \vspace{5pt}
 \caption{Comparison with baseline methods on SDD for $T_{obs}=8$ and $T_{pred}=12$, including SOTA PECNet.}
\label{tab:sdd}
\vspace{-10pt}
\end{center}
\end{table}

\begin{table}[tb!]
\begin{center}
 \begin{tabular}{c||c|c|c}
 \hline
  Method & minADE$_1$ & minFDE$_1$ & minADE$_5$ \\
  \hline
    MultiPath~\cite{chai2019multipath} & 28.32 & 58.38 & 17.51 \\
  \hline
    PCCSNet & \textbf{18.14} & \textbf{36.32} & \textbf{12.54} \\
  \hline
 \end{tabular}
 \vspace{5pt}
 \caption{Comparison with MultiPath on SDD for $T_{obs}=5$ and $T_{pred}=12$. minADE$_k$ and minFDE$_k$ measures the displacement error against the closest trajectory in top k samples.}
\label{tab:multipath}
\vspace{-10pt}
\end{center}
\end{table}

\begin{table}[tb!]
\begin{center}
 \begin{tabular}{c||c||c|c|c}
 \hline
  Method & KM w/o deep & KM & HAC & GMM \\
  \hline
    ADE & 0.24 & \textbf{0.21} & 0.21 & 0.21 \\
    FDE & 0.45 & \textbf{0.42} & 0.43 & 0.42 \\
  \hline
    Time/min & 0.6 & 0.7 & 18 & 90 \\
  \hline
 \end{tabular}
 \vspace{5pt}
 \caption{Comparison between different clustering methods on ETH/UCY Dataset. Results are the average of five sub-datasets.}
\label{tab:clustering methods}
\vspace{-10pt}
\end{center}
\end{table}

\begin{table}[tb!]
\begin{center}
 \begin{tabular}{c||c|c|c|c}
 \hline
  K & 100 & 200 & 500 & 1000 \\
  \hline
    ADE & 0.22 & \textbf{0.21} & 0.21 & 0.22 \\
    FDE & 0.44 & \textbf{0.42} & 0.43 & 0.45 \\
  \hline
 \end{tabular}
 \vspace{5pt}
 \caption{Comparison between different parameter $K$ in K-means on ETH/UCY Dataset. Results are the average of five sub-datasets.}
\label{tab:clustering K}
\vspace{-20pt}
\end{center}
\end{table}

\subsection{Quantitative Evaluation}

\vspace{5pt}\noindent\textbf{ETH/UCY.} Experimental results on ETH/UCY benchmark against competing methods are shown in Tab.~\ref{tab:eth}, including state-of-the-art STAR~\cite{yu2020spatio} and PECNet~\cite{mangalam2020not}. Note that the input information varies from different baselines, where P denotes historical path, S denotes social information and M denotes map information. To present the power of our framework more clearly, we only use historical paths as the information source. Results demonstrate that the performance of trajectory prediction is further elevated with our PCCSNet framework. We reach improvement of 19.2\% (0.05/0.26) and 20.8\% (0.11/0.53) on ADE and FDE in average respectively comparing with SOTA performance achieved by STAR. Notably, we achieve such improvement by using historical paths as the only input while STAR uses both paths and social information. 

Our method fails comparing with PECNet on FDE on some subsets. We attribute this to the major differences that our method is ADE-prioritized while PECNet is FDE-prioritized, which means that PECNet has a tendency to achieve a lower FDE than a lower ADE. Further, the social information is absent in our method.

\vspace{5pt}\noindent\textbf{SDD.} We also report the prediction performance on SDD dataset in Tab.~\ref{tab:sdd}. Comparing with the SOTA framework PECNet, we achieve remarkable improvement of 13.5\% (1.34/9.96) increase on ADE. There is a little FDE decline of 1.8\% (0.28/15.88). Considering the differences we have discussed above and the trade-off between huge ADE improvement and a minor FDE decline, we believe that our prediction results are promising. 

\begin{figure*}[tb!]
\centering
\vspace{-20pt}
\includegraphics[width=1.0\textwidth]{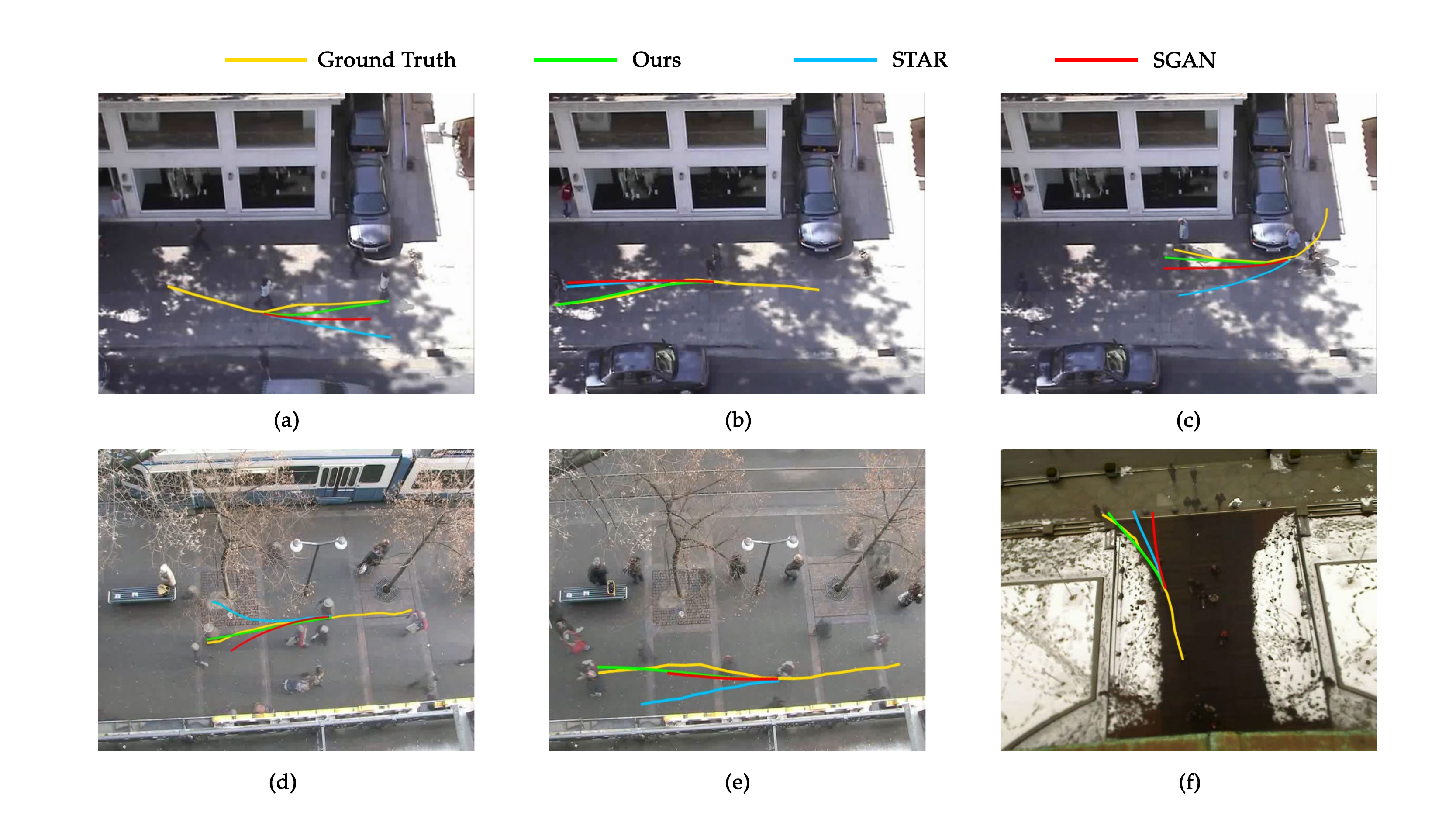}
\vspace{-20pt}
\caption{Accuracy analysis of PCCSNet where yellow lines indicate ground truth paths and our predicted results are in green. Predictions of two comparison models, STAR and SGAN, are denoted as blue and red. Examples are from different scenes including ZARA (a-c), HOTEL (d,e) and ETH (f).}
\label{fig:accuracy experiment}
\vspace{-0pt}
\end{figure*}

\subsection{Analysis}

\vspace{0pt}\noindent\textbf{Comparing with MultiPath~\cite{chai2019multipath}.} We compare with MultiPath in Tab.~\ref{tab:multipath}. Note that the frames of observation 5 in MultiPath is different from commonly used 8~\cite{sadeghian2019sophie,mangalam2020not}. Our method greatly outperforms MultiPath. 

\vspace{5pt}\noindent\textbf{Clustering on Deep Features.} Our clustering process is applied on deep features to explore better representations for different modalities. Comparison between k-means on trajectory coordinates (KM w/o deep) and deep features (KM) in Tab.~\ref{tab:clustering methods} proves that our deep clustering process can capture modality representations effectively.

\vspace{5pt}\noindent\textbf{Clustering Algorithm.} We compare the performance of three common clustering algorithms when they are used for constructing $\mathbb{M}$ in Eq.~\ref{eqn:modality set}, including K-means (KM), hierarchical agglomerative clustering (HAC) and Gaussian mixture model (GMM). Results in Tab.~\ref{tab:clustering methods} show that the simple K-means algorithm exhibits a huge advantage in speed with no decline in accuracy. Therefore, we use K-means as the clustering algorithm in our implementation. 

\vspace{5pt}\noindent\textbf{K in K-means.} We also study the effect of different $K$ in K-means algorithm on the results, shown in Tab.~\ref{tab:clustering K}. A larger $K$ can let each modality $M$ reveal more fine-grained representations, yet it may cause a reduction in classification accuracy. We assign $200$ for $K$ in our experiments for better overall performance. 

\vspace{5pt}\noindent\textbf{Weights for Clustering Distance.} For clustering distance as described in Eq.~\ref{eqn:L2 distance}, both historical and future representations are taken into consideration. We add weights $w_H$ and $w_F$ to balance them for clustering. According to experiments in Tab.~\ref{tab:clustering weights}, a proper weighting ratio near $1:1$ will get much better performance. If the weight of future representation grows, the performance drops a little. And if the weight of historical representation grows, the performance drops a lot. This reveals two facts: 1) Future features take a leading role in modality clustering. 2) Historical features play a supporting role to differentiate confusing situations for better clustering results. 

\begin{table}[tb!]
\begin{center}
 \begin{tabular}{c||c|c|c|c|c}
 \hline
  $(W_H:W_F)$ & 1:3 & 1:2 & 1:1 & 2:1 & 3:1 \\
  \hline
    ADE & 0.22 & 0.21 & \textbf{0.21} & 0.23 & 0.24 \\
    FDE & 0.43 & 0.42 & \textbf{0.42} & 0.47 & 0.49 \\
  \hline
 \end{tabular}
 \vspace{5pt}
 \caption{Comparison between different weight configurations for clustering distance in Eq.~\ref{eqn:L2 distance} on ETH/UCY Dataset. The results are the average of five sub-datasets.}
\label{tab:clustering weights}
\vspace{-10pt}
\end{center}
\end{table}

\begin{table}[tb!]
\begin{center}
 \begin{tabular}{c|c|c||c|c}
 \hline
  r/m & $\Delta v$ & $\Delta\theta$/$\pi$ & ADE & FDE \\
  \hline
    0 & 0 & 0 & 0.221 & 0.443 \\
  \hline
    \textbf{1} & \textbf{10\%} & \textbf{0.1} & \textbf{0.208} & \textbf{0.422}  \\
  \hline
    0.5 & 10\% & 0.1 & 0.212 & 0.426  \\
    2 & 10\% & 0.1 & 0.215 & 0.432  \\
  \hline
    1 & 5\% & 0.1 & 0.213 & 0.426  \\
    1 & 20\% & 0.1 & 0.216 & 0.438  \\
  \hline
    1 & 10\% & 0.05 & 0.212 & 0.427  \\
    1 & 10\% & 0.2 & 0.213 & 0.427  \\
  \hline
 \end{tabular}
 \vspace{5pt}
 \caption{Sensitive analysis on differet hyper-parameter configurations for Modality Loss on ETH/UCY Dataset. The results are the average of five sub-datasets.}
\label{tab:sensitive loss}
\vspace{-10pt}
\end{center}
\end{table}

\begin{table}[tb!]
\begin{center}
 \begin{tabular}{c||c|c|c}
 \hline
  Method & w/o synthesis & PCCSNet & $\Delta$ \\
  \hline
    ADE & 0.23 & \textbf{0.21} & 8.7\% \\
  \hline
    FDE & 0.45 & \textbf{0.42} & 6.7\% \\
  \hline
 \end{tabular}
 \vspace{5pt}
 \caption{Contribution of modality synthesis on ETH/UCY. Results are the average of five sub-datasets.}
\label{tab:synthesis}
\vspace{-20pt}
\end{center}
\end{table}

\vspace{5pt}\noindent\textbf{Analysis for Modality Loss.} We introduce Modality Loss in our framework to enhance the diversity and Tab.~\ref{tab:sensitive loss} shows a comprehensive analysis. After applying Modality Loss to enhance the diversity, our performance is further improved. Some hyper-parameters will decide the scale of restrictions of similar movements, including the radius of circle $r$, and thresholds for speed $\Delta v$ and angle $\Delta \theta$. When these parameters vary, the performance fluctuates, where speed constraint is more sensitive than others.

\vspace{5pt}\noindent\textbf{Contribution of Modality Synthesis.} Modality synthesis is proposed to optimize predictions at fine-grained level. Tab.~\ref{tab:synthesis} shows the synthesis process brings huge improvement in accuracy.

\begin{figure*}[tb!]
\centering
\vspace{-30pt}
\includegraphics[width=1.0\textwidth]{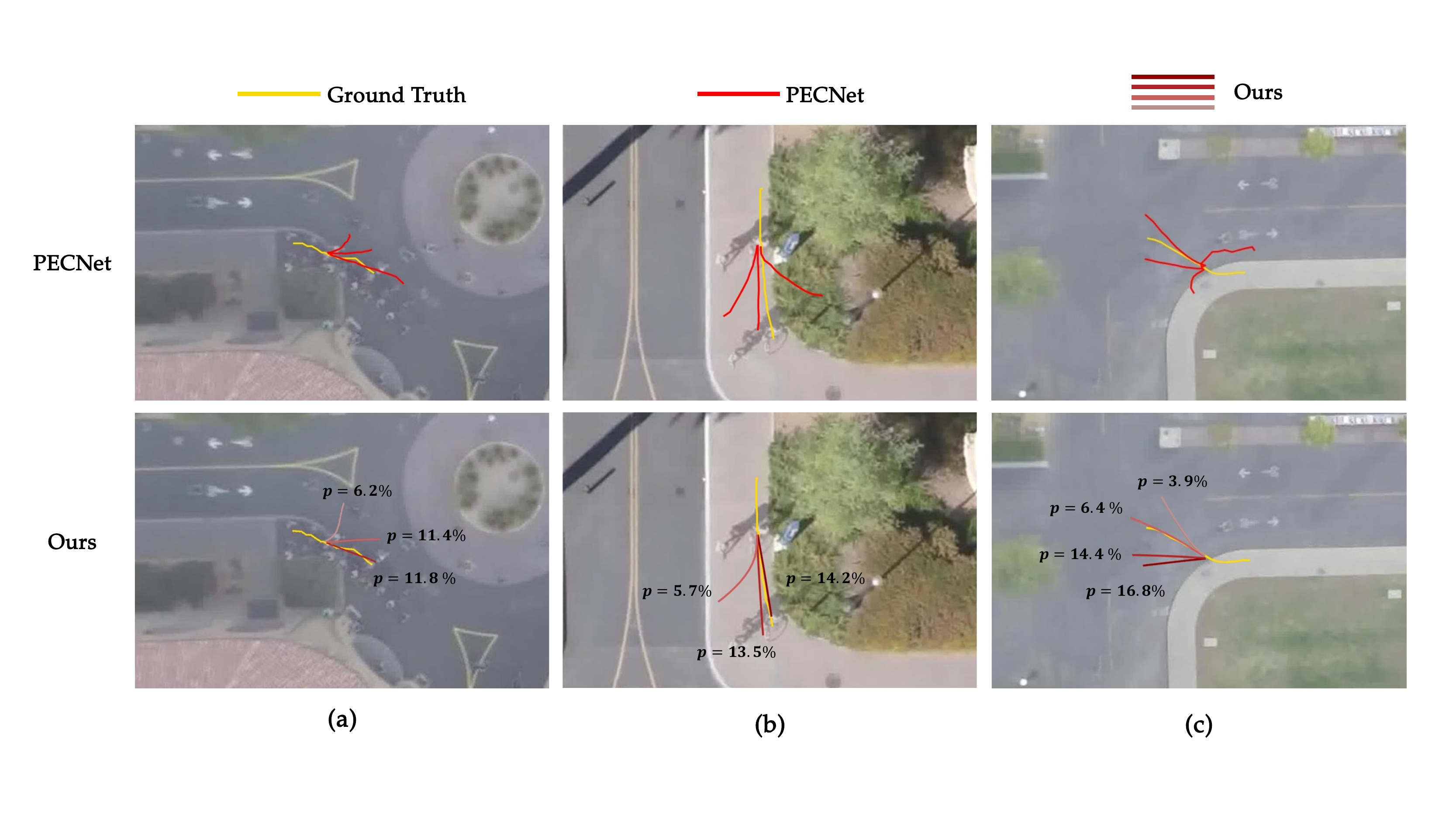}
\vspace{-30pt}
\caption{Illustration of our probabilistic multimodal predictions where yellow curves indicate ground truth paths and our predicted results are in red (darker refers to higher probability). PECNet is used for comparison and its results are denoted as the same dark red since each result it outputs has equal probability. The probabilities of our method are marked out while they take the same value of $5\%$ for PECNet (1/20). We only visualize some representative trajectories for a clear view.}
\label{fig:diversity experiment}
\vspace{-5pt}
\end{figure*}

\begin{figure}[tb!]
\centering
\includegraphics[trim=75 20 75 20, clip, width=1.0\linewidth]{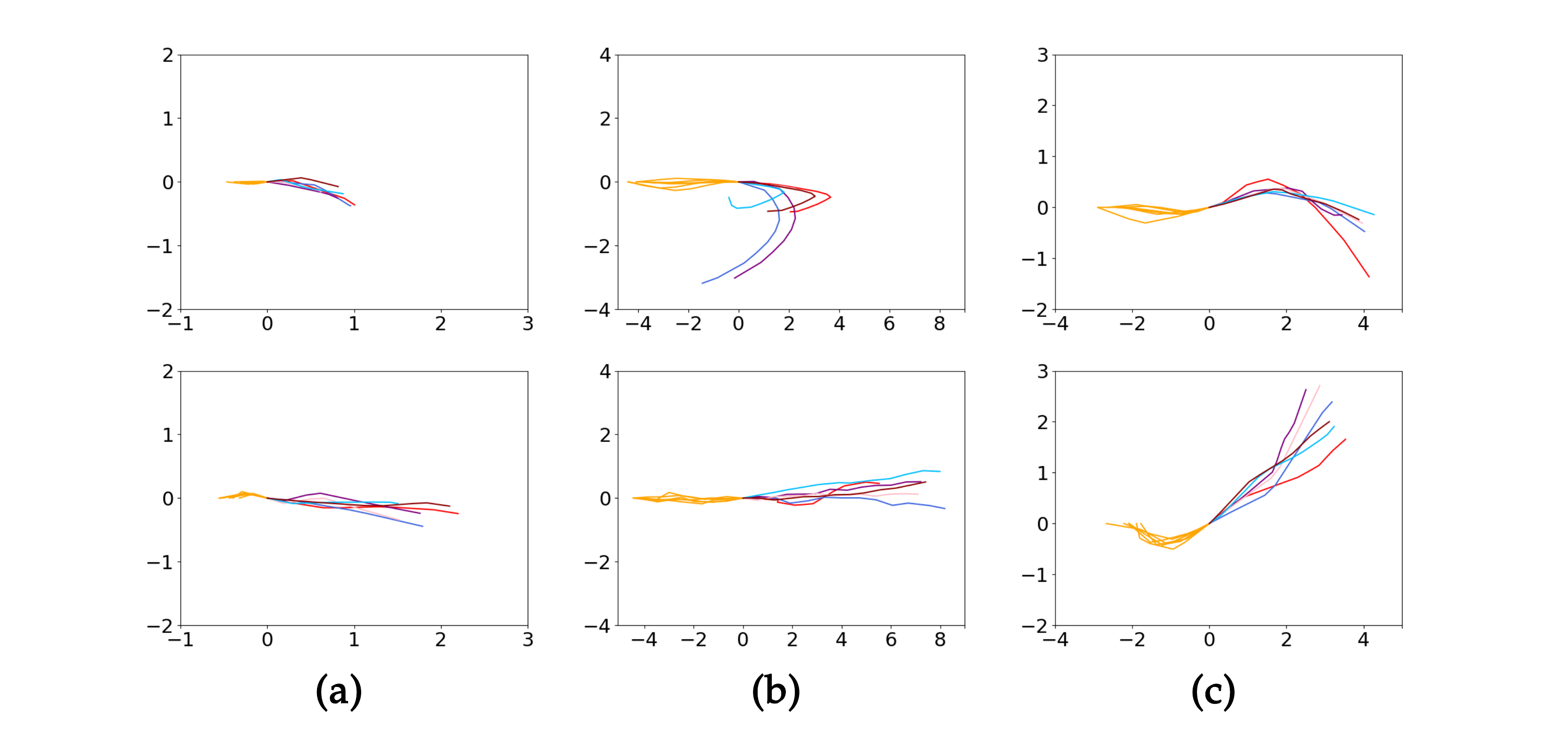}
\vspace{-5pt}
\caption{Visualization of clusters. Past trajectories are in yellow and future ones are in various colors for a clear view. Each column refers to a pair with similar past trajectories and different future modalities. Values of x,y-axis are coordinates in meters.}
\label{fig:clusters experiment}
\vspace{-10pt}
\end{figure}

\subsection{Qualitative Evaluation}

\vspace{0pt}\noindent\textbf{Accuracy Analysis.} We compare PCCSNet with other approaches on various challenging cases including turning and acceleration. Predicted results of the best modality are shown with different colors in Fig.~\ref{fig:accuracy experiment}, which illustrate that our model can capture much more accurate future modalities in terms of both speed and direction. Figure (c) and (f) are two typical cases of turning. Our model predicts more accurate angles with a proper speed. STAR gives a much smaller angle in both cases while SGAN predicts a wrong turning direction in (f). Another challenging case in figure (e) shows that even though PCCSNet ignores a jitter in the future, the predictions of speed, direction and destination are remarkable. Both STAR and SGAN fail to estimate neither the speed nor the destination.

\vspace{5pt}\noindent\textbf{Multimodality with Probability.} Fig.~\ref{fig:diversity experiment} demonstrates the probabilistic property of our proposed framework. In these three specific scenes, not only can our model give accurate predictions, probabilities of reasonable futures also tend to be much higher than average. Therefore, our model is less likely to predict trajectories that are improbable while maintaining the accuracy of the best-match. However, each prediction sampled by a generative model can only be interpreted as a average probability of $\frac{1}{k}$ when taking $k$ predictions. Thus, they lose probabilistic information for each modality, which is important to the following decision-making process.

\vspace{5pt}\noindent\textbf{Clustering Analysis.} To illustrate the effectiveness of the clustering algorithm in exploring different potential modalities, we visualize samples in some different clusters in Fig.~\ref{fig:clusters experiment}. Pair (a) indicates two types of possible futures that both come after a slow, linear past trajectory. The top one remains slow whereas the bottom one begins to accelerate. In pair (b), the past trajectories are still linear but faster. We show an $180^{\circ}$ turn and a straight path that may happen in the future. Pair (c) shows curved past trajectories unlike former ones. Corresponding future modalities are right and left turning. These cases give a strong proof that our clustering algorithm is sensitive enough to capture differences between potential modalities. Note that although the clustering is performed in a high dimensional space following Eq.~\ref{eqn:modality set}, the visualization is in 2D space for readability.

\section{Conclusion}
In this paper, we formulate the multimodal prediction framework into three steps of modality clustering, classification and synthesis to address major weaknesses in previous works and present a brand-new pipeline PCCSNet to solve it. Considering that the future are usually centralized around several different behaviors, we first cluster encoded historical and future representations to identify potential behavior modalities. A classifier then is trained to figure out the probability of occurrence for each modality given a historical path with a novel Modality Loss. Further, a modality synthesis mechanism is proposed to get fine-grained prediction results deterministically. Exhaustive experiments demonstrate the superiority of our elaborately designed framework in accuracy, diversity and reasonableness, even without introducing social and map information. 

{\small
\bibliographystyle{ieee_fullname}
\bibliography{egbib}
}

\end{document}